# ThUnderVolt: Enabling Aggressive Voltage Underscaling and Timing Error Resilience for Energy Efficient Deep Learning Accelerators

Jeff (Jun) Zhang
New York University
jeffjunzhang@nyu.edu

Kartheek Rangineni
New York University
kr2195@nyu.edu

Zahra Ghodsi
New York University
zg451@nyu.edu

Siddharth Garg
New York University
sg175@nyu.edu

## ABSTRACT

Hardware accelerators are being increasingly deployed to boost the performance and energy efficiency of deep neural network (DNN) inference. In this paper we propose Thundervolt, a new framework that enables aggressive *voltage underscaling* of high-performance DNN accelerators without compromising classification accuracy even in the presence of high timing error rates. Using post-synthesis timing simulations of a DNN accelerator modeled on the Google TPU, we show that Thundervolt enables between 34%-57% energy savings on state-of-the-art speech and image recognition benchmarks with less than 1% loss in classification accuracy and no performance loss. Further, we show that Thundervolt is synergistic with and can further increase the energy efficiency of commonly used run-time DNN pruning techniques like Zero-Skip.

## 1 INTRODUCTION

Deep neural networks (DNN) have achieved or surpassed state-of-the-art results in a wide range of machine learning applications, from image, video and text classification to speech recognition and language translation [37]. DNNs contain multiple layers of computation, where each layer multiplies inputs from the previous layer with a matrix of weights followed by a non-linear transformation such as a sigmoid or linear rectification. State-of-the-art DNNs have *millions* of parameters requiring *large* matrix multiplications (or convolutions) and are thus computationally challenging workloads. Consequently, there have been several efforts from academia and industry on the design of special-purpose hardware accelerators for both DNN training and DNN inference [6, 7, 11, 21, 30, 35]. This paper focuses specifically on hardware accelerators for DNN inference.

While several different architectures for DNN acceleration have been proposed, *systolic array* based implementations are among the most promising [21, 25, 40]. Systolic arrays are 2-D grids of homogeneous processing elements (PE) that perform only nearest neighbour communication, thus obviating the need for input buffering and complex routing. Furthermore, systolic arrays enhance energy efficiency by amortizing frequent (and power-hungry) reads of DNN weights and activations over multiple compute operations. The Tensor Processing Unit (TPU), an accelerator for DNN inference that is currently deployed in Google datacenters, uses a systolic array with a $256 \times 256$ grid of multiply-and-accumulate (MAC) units at its core and provides $30\times$-$80\times$ greater performance/Watt than CPU and GPU based servers at a raw throughput of 90 TOPS/second [21].

Voltage underscaling based timing speculation is a promising technique that enables the energy consumption of digital circuits to be reduced *beyond* that achievable by conventional worst-case design methodologies [12]. The idea is based on the observation that worst-case timing critical paths in digital logic are *rarely* exercised, making it possible to run digital circuits at supply voltages lower than the nominal voltage if timing errors can be dealt with. Two broad techniques are proposed in literature to cope with timing errors: (i) *timing error detection and recovery* (TED) [9, 12–14]; and (ii) *timing error propagation* (TEP) [20, 23, 28, 42, 43]. TED detects timing errors (using Razor flip-flops [12], for instance) and recovers by safely re-executing the offending input. TEP, on the other hand, allows errors to propagate to subsequent logic stages in the expectation that the algorithm itself is error resilient. Unfortunately, as discussed below and in Section 2, TED and TEP only provide limited voltage underscaling opportunities for high-performance DNN accelerators.

The challenge with applying TED to DNN accelerators with thousands of parallel MAC units is that the global timing error rate (likelihood of *at least* one MAC experience a timing error) grows dramatically with the number of MAC units — a TPU-like array with roughly 65K MACs, for instance, has a 50% global timing error rate even if each MAC unit only experiences $10^{-5}$ timing errors/clock cycle. Even assuming an ideal single-cycle recovery penalty, a 50% global timing error rate imposes significant performance (and energy) overheads (see Section 2 for more detailed evaluation). On the other hand, although TEP does not have a performance penalty, prior work [20] and our own empirical evaluations in Section 4.2 show that simply allowing timing-induced errors to propagate results in significant drops in classification accuracy even at timing error rates as low as 0.1%.

In this paper, we propose **Thundervolt**, a framework that enables *aggressive* voltage underscaling of DNN accelerators without compromising performance and with negligible (< 1%) impact on classification accuracy. Specifically, we make the following novel contributions:

- We propose **TE-Drop**, a new timing error recovery technique for DNN accelerators. TE-Drop detects timing errors using Razor flip-flops (like TED), but avoids the performance impact of re-execution by *dropping* the MAC operation subsequent to an erroneous MAC and using the extra clock cycle to correctly compute the erroneous MAC's result. Empirically, we show that TE-Drop enables execution at at timing error rates greater than 10% without compromising accuracy.
- Second, we show empirically that the timing error rate of state-of-the-art DNNs varies significantly across layers. Motivated

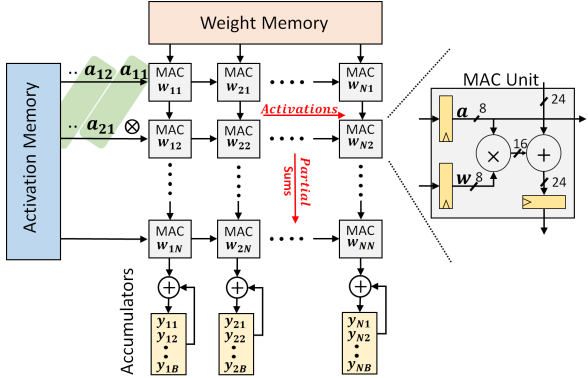

Figure 1: Architecture of a systolic array based DNN accelerator that serves as a baseline for Thundervolt.

by this observation, we propose a **dynamic per-layer voltage underscaling** scheme that determines the voltage underscaling ratio of each layer to balance timing error rates across layers of the DNN. Empirically, the proposed scheme significantly outperforms an alternative in which each layer is under-scaled by the same amount.

- We show that Thundervolt can be applied **synergistically with Zero-Skip**, an energy reduction technique employed in several DNN accelerators [1, 7, 22, 29, 32]. Notably, while Zero-Skip reduces the energy of MACs with zero inputs (either weight or activation), Thundervolt reduces the energy of all remaining non-zero MAC operations by executing them at an underscaled voltage.
- Empirical evaluations of Thundervolt based on detailed post-synthesis timing simulations show that Thundervolt enables between 34.36%-57.75% energy savings over a baseline DNN accelerator modeled on the Google TPU enhanced with Zero-Skip optimization. These savings are achieved with no performance overhead and at most a 1% reduction in classification accuracy.

To the best of our knowledge, ours is the *first* paper to investigate the use voltage underscaling based timing speculation for high-performance, highly-parallel DNN accelerators like the Google TPU.

## 2 BACKGROUND AND MOTIVATION

In this section, we describe the relevant background on hardware acceleration for DNNs, and motivate Thundervolt by discussing why existing timing speculation mechanisms such as TED and TEP can only provide limited energy savings for high-performance DNN accelerators.

### 2.1 Hardware Accelerators DNNs

We begin by mathematically describing how a DNN performs inference (i.e., predictions) on a *batch* of $B$ inputs. A DNN consists of $L$ stacked layers of computation. Layer $i$ has $N^i$ *neurons* whose outputs are referred to as *activations*, represented by an matrix $A^i \in \mathbb{R}^{N^i \times B}$, in which each column corresponds to an input from the batch. Layer $i$ multiplies its activation matrix with weight matrix $W^i \in \mathbb{R}^{N^{i+1} \times N^i}$ and adds a bias $b^i \in \mathbb{R}^{N^{i+1}}$ to yield $Y^i = W^i A^i + b^i$, where $Y^i \in \mathbb{R}^{N^{i+1} \times B}$. The matrix multiplication outputs are passed through an element-wise non-linearity $\phi(.)$, such as the tanh, sigmoid and rectified linear (ReLU) non-linearities [37], to yield the activations for the next layer, i.e., $A^{i+1} = \phi(Y^i)$. Note that DNNs are generalizations of *convolutional neural networks* (CNN) in which the matrix multiplication is replaced with convolutions; the proposed techniques apply to CNNs as well but a detailed description of CNN execution is omitted for lack of space.

DNN accelerators aim to speed-up the most computationally expensive operation in DNN execution, i.e., computing the matrix multiplication $W \times A$ (the superscript indicating the layer is dropped for simplicity). Figure 1 shows a block-diagram of an exemplar DNN accelerator modeled closely on the TPU that contains, at its heart, a systolic array with an $N \times N$ grid of MAC units that accelerates matrix multiplication. Assume that $w_{ij}$ ($a_{ij}$) is the element in the $i^{th}$ row and $j^{th}$ column of weight (activation) matrix $W$ ($A$). The operation of the systolic array can be explained as follows.

First, weights are pre-loaded into array and remain stationary throughout a block of computation. We refer to the MAC unit that stores $w_{ij}$ as $\text{MAC}_{ij}$. Next, activations stream in from the activation memory, one activation per row per clock cycle, and move from left to right. The activation streams are purposefully *staggered* by one clock cycle. $\text{MAC}_{11}$ computes $w_{11}a_{11}$ in the first clock cycle, $\text{MAC}_{12}$ unit adds $w_{12}a_{21}$ to $\text{MAC}_{11}$'s product in the next clock cycle, and so on. In clock cycle $N$, the $\text{MAC}_{1N}$ unit outputs $y_{11} = \sum_i w_{1i}a_{i1}$, which is the first element of the output matrix. $\text{MAC}_{1N}$ proceeds to output $y_{12}, \ldots, y_{1B}$ in subsequent clock cycles.

The second column receives the same stream of inputs as the first column, but delayed by one clock cycle. This column outputs $y_{21}, y_{22}, \ldots, y_{2B}$. In this manner, a batch of $B$ inputs is multiplied by an $N \times N$ weight matrix in $2N + B$ clock cycles.

If the size of matrix-matrix multiplication is larger than the systolic array, the operation is performed in several blocks. The result of each block is stored in accumulators and is added to the result of the next block. After matrix-matrix multiplication is complete, the activation function is applied to the results stored in accumulators using a separate layer of logic. For ReLU non-linearities, only a single comparator per accumulator is required and therefore the activation layer constitutes less than 0.3% of the systolic array's total area. Note that convolutional layers can also use the same DNN architecture by loading weights and activation values into the systolic array with an appropriate schedule.

### 2.2 Timing Speculation for DNN Accelerators

We now discuss the limitations of conventional timing speculation schemes, i.e., TED and TEP, in the context of DNN accelerators.

**Timing Error Detection and Recovery (TED)** First proposed by Ernst et al. [12], TED uses so-called Razor flip-flops to sample the output of combinational logic twice: once using the regular clock and again using a delayed clock. If the two samples are different, a timing error is flagged and the input is safely re-executed at a reduced clock frequency by gating the subsequent clock edge.

An ideal TED scheme detects if *any* flip-flop in the design has a timing error, and if so, distributes a clock gating signal to each flip-flop, all within one clock cycle. We refer to this technique as *global* TED. As the authors note [12], global TED is impractical for even moderately sized designs. Consequently, several *distributed* TED schemes that use pipelined error detection and recovery logic have been proposed [9, 13], but incur higher performance overhead than global TED.

Figure 2a illustrates the limitations of global TED on a 256 × 256 MAC array; note from the figure that even though the local timing error rate (i.e., timing errors per MAC operation) increases only gradually with voltage underscaling, the global timing error rate shoots up when the voltage is underscaled below $r = 0.9$ of the nominal voltage ($r$ is referred to as the voltage underscaling ratio). Therefore, even optimistically assuming that single-cycle error detection and recovery is feasible, voltage underscaling below $r = 0.9$ results in *increased* energy consumption. A more realistic five cycle error recovery penalty further increases the energy costs of global TED.

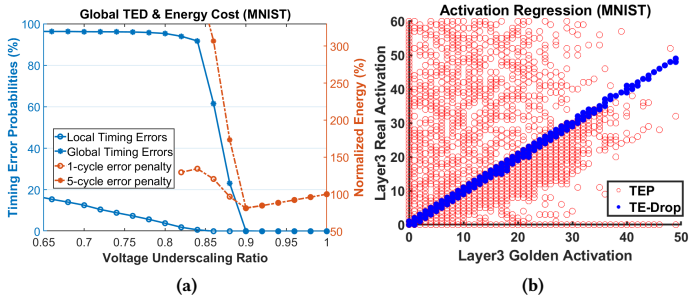

Figure 2: (a)Timing error probabilities versus voltage underscaling ratio, and the corresponding energy cost for global TED. (b) TEP versus the error-free outputs of the network.

While we omit a detailed discussion of distributed TED schemes for brevity, we note that a timing error in any MAC unit introduces a bubble that is propagated downstream (towards the right and downwards in Figure 1) and a stall signal that propagates upstream (towards the left and upwards in Figure 1). When the stall signal eventually reaches the left-most column, it causes *all* of the array's $N$ parallel inputs to skip one clock cycle. Note that to maintain synchrony in the systolic array, it is important that all inputs skip a cycle. Consequently, distributed TED schemes perform no better than global TED.

**Timing Error Propagation (TEP)** An alternative approach is to exploit algorithmic noise tolerance and simply allow timing errors to propagate to subsequent stages of computation instead of re-executing inputs that cause errors [18, 20, 42]. Recent work [20] has demonstrated that TEP causes the classification accuracy of DNNs to drop sharply for timing error rates as low as 0.1%. Our empirical evaluations of TEP in Section 4.2 reach the same conclusion.

The sensitivity of TEP to timing errors is because the errors frequently flip high-order bits of MAC outputs (i.e., the partial sums), resulting in large computational errors even at low timing error rates. Figure 2b plots the error-free (golden) vs. the TEP activations for Layer 3 of a MNIST digit recognition network executing at a voltage underscaling ratio of $r = 0.85$. The timing error rate per MAC at $r = 0.85$ is only 0.01%; yet the root mean square error between the TEP and golden activations is as high as $33.9x$ the average value of golden activations.

## 3 THUNDERVOLT DESIGN

We now describe in detail the new architectural innovations that we propose as part of of Thundervolt.

### 3.1 TE-Drop

At the heart of Thundervolt is TE-Drop, a new technique to deal with timing errors in MAC units. Like TED, TE-Drop instruments Thundervolt with Razor flip-flops to detect timing errors, but recovers from timing errors *without* re-executing erroneous MAC operations.

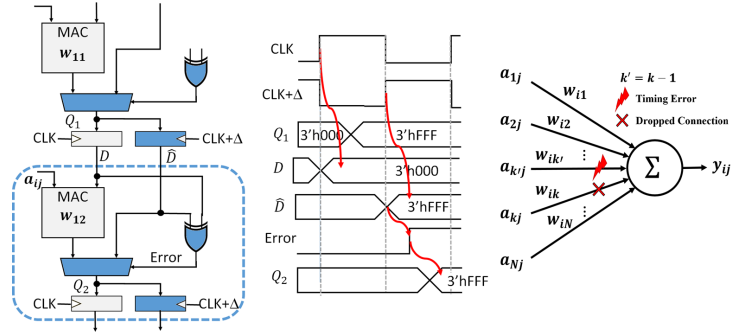

Figure 3: A block-level diagram illustrating TE-Drop and its impact of timing errors on the computation of a neuron.

TE-Drop builds on the observation that the distribution of weights in DNNs (and CNNs) is biased towards small values [16] (we also validate this observation in our own experiments); hence, the contribution of each individual MAC operation to the output of a neuron is small.

When a MAC incurs a timing error, TE-Drop steals the next clock cycle from its successor MAC to correctly finish its own update to the partial sum, and *bypass* (or drops) the successor MAC's update. As shown in Figure 3, TE-Drop requires minimal hardware changes. In addition to Razor flip-flops, TE-Drop adds a multiplexor (MUX) controlled by the error signal from the prior MAC unit. If the previous MAC unit incurs an error, the MUX forwards the previous MAC's correctly computed partial sum (obtained from Razor's shadow flip-flop) to the next MAC unit; if not, the current MAC unit updates the partial sum as it normally would and forwards to the next MAC.

Figure 3 illustrates TE-Drop's operation with a timing diagram. As in prior work, we assume that the shadow clock is delayed by 50% of the clock period, after which the error signal and correct partial sum from the prior MAC become available. Note that, although not shown in Figure 3, the error signal is obtained by OR-ing the error outputs of each of individual Razor flip-flop at the output of the MAC. This requires at worst a 24-input OR gate, although, as we note in Section 4.1, only a fraction of the MAC units outputs need to be protected using Razor flip-flops. For correct operation, the error signal computation and propagation through the MUX must complete within the remaining half clock cycle. This constraint is easily met; in our implementation, error signal computation and MUX propagation consume only 16.7% and 2% of the clock period.

The effect of TE-Drop on the DNN execution can be captured mathematically as follows. Let $e_{ijk}$ be a binary variable that is zero if the MAC operation that multiplies weight $w_{ik}$ with activation $a_{kj}$ is erroneous and one otherwise. The systolic array augmented with TE-Drop computes the output $y_{ij}$ as follows: $y_{ij} = \sum_{k=1}^{N} w_{ik} a_{kj} e_{ijk'}$, where $k' = k - 1$. Equivalently, as illustrated in Figure 3, TE-Drop can be viewed as "randomly" dropping a fraction of connection between input and output neurons, where the randomness arises from timing errors. Interestingly, randomly dropping neurons and connections during training are commonly used techniques, referred to as dropout [19] and dropconnect [39], respectively, that prevent overfitting and help reduce generalization error.

### 3.2 Per-layer Voltage Underscaling

Our baseline accelerator computes on each layer of the DNN iteratively. Each layer can take thousands of clock cycles to execute, depending on its size. We have so far assumed that the voltage underscaling ratio, $r$, is constant across layers. However, empirical evidence suggests that it might be beneficial to dynamically adjust the extent of of voltage underscaling from one layer to the next.

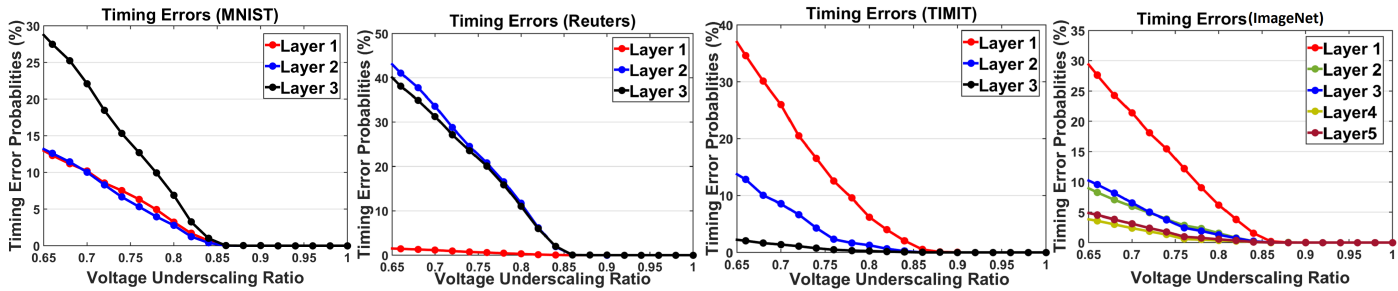

Figure 4: Timing error probabilities for each layer of (a) MNIST, (b) Reuters, (c) TIMIT, and (d) ImageNet.

Figure 4 plots the timing error rate for each layer of four DNN architectures (more details about each DNN can be found in Section 4.1) as a function of the voltage underscaling ratio. In all four cases, we find that the timing error rate varies significantly from one layer to the next. For instance, at the lowest voltage underscaling ratio of $r = 0.65$, the error rates for layers with the highest and lowest timing error rates differs by between 2× to 20×. Further, we find that for the two smaller networks, MNIST and Reuters, the timing error rate is highest for deeper layers in the network, while for the two larger networks, TIMIT and AlexNet, the timing error rate decreases with depth.

Using the same voltage underscaling ratio for each layer might limit the potential for some layers to run at low voltages because the high error rate for another layer has an unacceptable impact on classification accuracy. Hence, we propose per-layer voltage underscaling scheme that aims to equally distribute a total timing error budget, $p_{total}$, equally among layers. Our approach is described in Algorithm 1.

Algorithm 1 takes as input the timing error profile as a function of the voltage underscaling ratio for each layer in the DNN (characterized by the function $p_i(r)$), a net timing error budget $p_{total}$, and the minimum allowed voltage underscaling ratio $r_{min}$, and outputs the optimal voltage underscaling ratio $r_i^*$ for each layer $i \in [1, L]$.

The timing error rates per layer can be obtained using a validation dataset via timing simulations, or, practically, from production chips via the online error tracking logic shown in Figure 5. The error tracking logic sums up the single-bit error signals generated by MAC units along the array's columns. The total error of each column is then summed along the bottom row to obtain the total error for an entire batch of execution. The proposed error tracking logic has low area overheads (less than 10%), and has significant timing slack. Further, the circuitry does not incur in-field power overhead (on test data) once the optimal underscaling ratios are determined using Algorithm1. Note that in the presence of process variations, the timing error profile of each chip may be different and is characterized separately.

Algorithm 1 first sorts the layers in ascending order of timing error rate, using the timing error of each layer at the minimum voltage underscaling ratio. Next, the algorithm loops through layers in sorted order and in each iteration computes a target error rate for the layer by dividing the remaining error budget (initialized to $p_{total}$) by the remaining number of layers. The voltage underscaling ratio for the

**Algorithm 1:** Per-layer voltage underscaling algorithm.

**Data:** $L, p_{total}, r_{min}, p_i(r) \; \forall i \in [1, L]$
**Result:** $r_i^* \; \forall i \in [1, L]$

1 $p_{remain} = p_{total}$;
2 Sort layers in ascending order of $p_i(r_{min})$;
3 Store sorted order in vector $idx$;
4 **for** $i \in [1, L]$ **do**
5 $\quad j = idx(i)$;
6 $\quad p_{target} = \frac{p_{remain}}{L-i+1}$;
7 $\quad r_j^* = \arg\min_{r: p_j(r) \le p_{target}} (p_j(r))$;
8 $\quad p_{remain} = p_{total} - p_j(r_j^*)$;
9 **end**

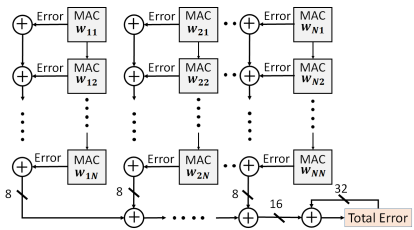

Figure 5: Online distributed error tracking circuitry.

current layer is set to the ratio that achieves the largest error probability that is less than or equal to the target. Finally, the current layer's error probability at this ratio is subtracted from the remaining budget. The rationale for looping through the layers in ascending order of timing probability is to ensure that low error probability layers do not get assigned budgets they cannot fully utilize.

## 4 EMPIRICAL EVALUATION

### 4.1 Experimental Setup

Table 1: Details of the DNN benchmarks used and accuracy of 32-bit and 8-bit implementations for the entire test set.

| Benchmarks | | Accuracy (%) | |
|---|---|---|---|
| Name | Architecture | 32-bit | 8-bit |
| MNIST | L1-L4 (FC): 784 × 256 × 256 × 256 × 10 | 98.15 | 98.15 |
| Reuters [5] | L1-L4 (FC): 2048 × 256 × 256 × 256 × 52 | 91.32 | 91.32 |
| TIMIT [4] | L1-L4 (FC): 1845 × 2000 × 2000 × 2000 × 183 | 74.13 | 73.91 |
| ImageNet [10] | L1-L2 (Conv): (224, 224, 3) × (27, 27, 64) × (13, 13, 192) L3-L5 (Conv): (13, 13, 384) × (13, 13, 256) × (6, 6, 256) L6-L8 (FC): 4096 × 4096 × 1000 | Top 5 79.06 | Top 5 76.33 |

**Benchmarks.** We evaluate Thundervolt on four benchmarks, MNIST digit classification, Reuters text categorization [5], TIMIT speech recognition [4] and image recognition using the ImageNet dataset [10]. The first three benchmarks use multi-layer perceptrons (MLPs) trained from scratch, while for ImageNet, we used the pre-trained AlexNet CNN from PyTorch based on [24]. We note that MLPs account more than 66% of Google's datacenter workload [21]. Table 1 shows the parameters of the four DNN architectures.

**Thundervolt Hardware Parameters** Our baseline DNN accelerator is modeled around the Google TPU (see Figure 1) and contains $256 \times 256$ MACs with 8-bit weights and activations and a 24-bit output. We implemented a prototype of the baseline accelerator in synthesizable Verilog, and synthesized it with OSU FreePDK 45 $nm$ standard cell library using the Cadence Genus synthesis tool. Note that although Thundervolt performs voltage underscaling on the accelerator's systolic array only, we estimate that the array consumes roughly 80% of chip's power, ignoring peripherals. Further, we determine that only 14 out of 40 flip-flops in the MAC unit needed to be protected by Razor, which imposes a roughly 3.35% power penalty, consistent with [12].

**Simulation Methodology** We scheduled the benchmark DNNs on the systolic array using the mapping described in [34]. The three MLPs are run with a batch size of 256 inputs, while for ImageNet each batch has a single image, corresponding to an *effective* batch size of 3025 for the input convolutional layer and 169 for the output convolutional layer. As in [2, 29, 31], Thundervolt is only used to accelerate ImageNet's convolutional layers, which make up more than 90% of its computational cost.

We model timing errors via detailed post-synthesis gate-level timing simulations using ModelSim. For the three MLP benchmarks, we simulate the *entire* $256 \times 256$ systolic array. However, simulating even a single ImageNet input takes 8 hours on a 40-core Intel Xeon server. Thus, for the ImageNet CNN we obtain error probabilities from detailed timing simulations of 32 randomly selected columns, and inject timing errors in the remaining columns [20] based on these probabilities. For each DNN, we executed $7 \times 256$ inputs, of which 256 are used as validation data, while the remaining are used for test.

### 4.2 Experimental Results

**Energy vs. Accuracy Trade-offs** We begin by evaluating Thundervolt's energy vs. accuracy trade-offs on the validation dataset in Figure 6. We report results for two versions of Thundervolt: **ThVolt-Static** where each layer executes at the same voltage underscaling ratio, and **ThVolt-Dynamic** that performs per-layer underscaling based on Algorithm 1. Also shown, for comparison, is energy vs. accuracy data for a baseline TEP scheme [20]. The results for ThVolt-Static are obtained by sweeping voltage underscaling ratios, and those for ThVolt-Dynamic are obtained by sweeping the total timing error budget, $p_{total}$, in Algorithm 1.

Several observations can be made from Figure 6. First, note that for all four benchmarks, the accuracy of TEP falls sharply beyond an energy savings of 20% (corresponding to a voltage underscaling ration of $r = 0.9$). From Figure 2a, it can be seen that this is the point at which timing errors first start appearing, demonstrating the poor timing error resilience of TEP. In contrast, both ThVolt-Static and ThVolt-Dynamic continue to provide high accuracy even with aggressive voltage underscaling. Second, ThVolt-Dynamic consistently outperforms ThVolt-Static, demonstrating the benefits of balancing error probabilities across layers. Finally, of the two large benchmarks, TIMIT and ImageNet, the former is more resilient to timing errors — we believe this is in part because ImageNet has 1000 output classes while TIMIT has only 138. Hence, for ImageNet, the distance between the prediction probabilities for the top two classes will be lower, making it more susceptible to noise.

**Energy Savings on Test Data** The validation data in Figure 6 is used to select the lowest energy design point for each benchmark where the loss in classification accuracy compared to nominal is less than 1%. The selected voltage underscaling ratios and corresponding timing error rates are marked in Figure 6. Note that the reported timing error rates are per MAC operation; the corresponding global error rates at these design points are much higher. In fact, we find that over all four benchmarks, even an ideal TED implementation achieves at most 20% energy savings.

In Figure 7, we use the test dataset to compare the energy consumption of ThVolt-Dynamic with (a) the baseline TPU (base); (b) Zero-Skip [1, 7, 22, 29, 32], that saves energy at run-time by skipping over all MAC operations with zero weights and/or activations, and (c) Zero-Skip+ThVolt-Dynamic, that applies ThVolt-Dynamic on *top* of Zero-Skip. Several obervations can be made.

First, compared to the baseline, ThVolt-Dynamic yields energy savings ranging from 36.3% on ImageNet, 50% on TIMIT and 56% on MNIST and Reuters. Interestingly, applying ThVolt-Dynamic on top of Zero-Skip (i.e., Zero-Skip+ThVolt-Dynamic) provides roughly the same energy savings over Zero-Skip alone, demonstrating that Thundervolt is indeed *synergistic* with the Zero-Skip optimization. Finally, Zero-Skip+ThVolt-Dynamic, the best performing scheme reduces energy consumption over the baseline by 73.27%-89.49%. We repeated these experiments by injecting $\sigma = 5\%$ process-induced variations over 50 chip samples and note that the *average* energy savings remain the same. However, the distribution of energy savings is not shown for lack of space.

## 5 RELATED WORK

Thundervolt builds on a considerable body of literature on timing speculation based TED techniques [12, 13]. Watmough et al. [43] recently implemented TED on a small, ultra-low power DNN accelerator with only 16 parallel MAC units. However, as discussed in Section 2 and Section 4.2, these baseline TED techniques provide limited energy saving potential for highly-parallel, high performance DNN accelerators because of high global timing error rates. Techniques like time borrowing [8, 43] and timing-error aware synthesis [14, 27, 38] reduce baseline timing error rates and can be leveraged by Thundervolt to enable more aggressive underscaling.

Approximate computing techniques have also been leveraged in the design of energy-efficient DNN accelerators including techniques that use approximate multipliers [36] and reduced precision arithmetic [15, 17, 26, 32]. Here, the approximations are *by design*, while Thundervolt approximates due to timing errors. Thundervolt can potentially be applied on top of these techniques to exploit timing slack in approximate circuits. A related line of research uses stochastic computing for energy efficient DNN accelerators [3, 33]; however, stochastic logic based designs have yet not been adopted in commercial chips.

Another commonly used technique for energy efficient DNN hardware prunes neurons and connections at *design-time* [16, 41, 44], thus reducing the size of the DNN. While we have evaluated Thundervolt using standard architectural parameters reported in prior work, we note that Thundervolt can be applied on a pruned DNN as well. A quantitative evaluation of Thundervolt on compressed DNN models is demarcated as future work. Another common technique to enhance DNN efficiency is *run-time* pruning [1, 7, 22, 29, 32]; we show in

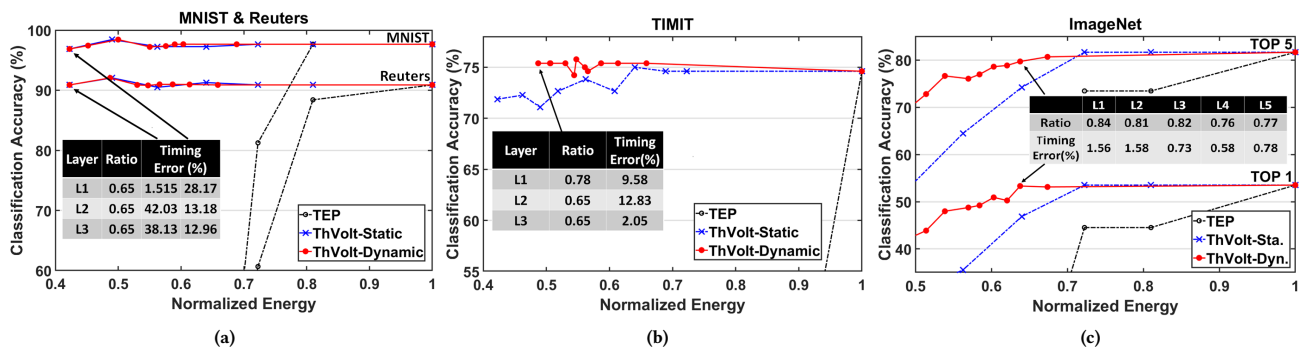

Figure 6: Validation Data: Accuracy energy tradeoff using Thundervolt.

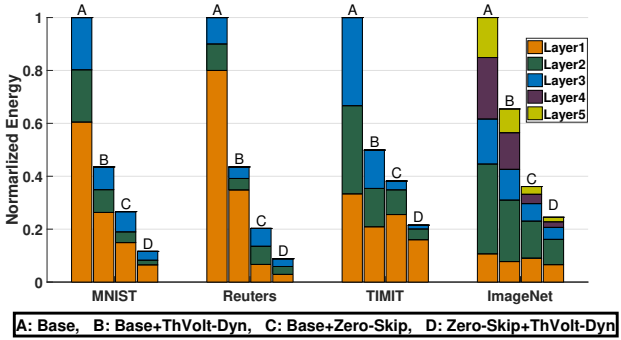

Figure 7: Layer-wise energy reduction techniques compared to the baseline. For all cases, the classification accuracy loss is within 1% of nominal, and no performance loss.

Section 4.2 that Thundervolt is synergistic with Zero-Skipping, a commonly used instantiation of run-time pruning.

## 6 CONCLUSION AND FUTURE WORK

In this paper, we have presented and evaluated Thundervolt, a new framework to enable aggressive voltage underscaling of high-performance DNN accelerators. Thundervolt deals with timing errors using a novel TE-Drop mechanism that incurs no performance penalty while providing robust classification accuracy. Further, Thundervolt performs per-layer dynamic voltage underscaling and can be synergistically applied with dynamic DNN pruning techniques to further reduce energy consumption. Our experiments demonstrate significant reductions in energy consumption at no performance cost and less than 1% accuracy loss. As future work, we would like to explore the synergy between Thundervolt and techniques such as time borrowing and approximate or reduced precision MAC units. Further, given the appealing analogy between TE-Drop and neural network regularization techniques like dropconnect, we would like to explore the use of Thundervolt in accelerating DNN training.